\documentclass[10pt,twocolumn,letterpaper]{article}

\usepackage[pagenumbers]{cvpr} %

\usepackage[dvipsnames]{xcolor}

\definecolor{cvprblue}{rgb}{0.21,0.49,0.74}
\usepackage[pagebackref,breaklinks,colorlinks,citecolor=cvprblue]{hyperref}
\usepackage{balance}

\def\paperID{*****} %
\def\confName{CVPR}
\def\confYear{2024}

\title{
\vspace{5px}
OpenSUN3D: 1$\mathrm{^{st}}$ Workshop Challenge on Open-Vocabulary\\3D Scene Understanding}

\author{
\emph{Workshop organizers:}\\[13px]
    Francis Engelmann\textsuperscript{1,2},
    Ayca Takmaz\textsuperscript{1},
    Jonas Schult\textsuperscript{3},
    Elisabette Fedele\textsuperscript{1},
    Johanna Wald\textsuperscript{2},
    Songyou\\Peng\textsuperscript{1},
    Xi Wang\textsuperscript{1},
    Or Litany\textsuperscript{4,5},
    Siyu Tang\textsuperscript{1},
    Federico Tombari\textsuperscript{2,6},
    Marc Pollefeys\textsuperscript{1,7},
    Leonidas Guibas\textsuperscript{8}\\
{\small
$^1$ETH Zurich \hspace{10px}
$^2$Google \hspace{10px}
$^3$RWTH Aachen \hspace{10px}
$^4$Technion \hspace{10px}
$^5$NVIDIA \hspace{10px}
$^6$TUM \hspace{10px}
$^7$Microsoft \hspace{10px}
$^8$Stanford University}\\\\[13px]
\emph{Challenge winners:}\\[13px]
    Hongbo Tian,
    Chunjie Wang,
    Xiaosheng Yan,
    Bingwen Wang,
    Xuanyang Zhang,
    Xiao Liu \\
    Phuc Nguyen,
    Khoi Nguyen,
    Anh Tran,
    Cuong Pham\\
    Zhening Huang,
    Xiaoyang Wu,
    Xi Chen,
    Hengshuang Zhao,
    Lei Zhu,
    Joan Lasenby\\
}

\begin{document}
\maketitle

\begin{abstract}
This report provides an overview of the challenge hosted at the OpenSUN3D Workshop on Open-Vocabulary 3D Scene Understanding held in conjunction with ICCV 2023.
The goal of this workshop series is to provide a platform for exploration and discussion of open-vocabulary 3D scene understanding tasks, including but not limited to segmentation, detection and mapping.
We provide an overview of the challenge hosted at the workshop, present the challenge dataset, the evaluation methodology, and brief descriptions of the winning methods.
Additional details are available on the \href{https://opensun3d.github.io/index_iccv23.html}{OpenSUN3D} workshop website.

\end{abstract}

\section{Introduction}
\label{sec:intro}

The ability to perceive, understand and interact with arbitrary 3D environments is a long-standing research goal with applications in AR/VR, robotics, health and industry.
Current 3D scene understanding methods are largely limited to recognizing a \emph{closed-set} of pre-defined object classes.
Recently, large visual-language models (VLMs), such as CLIP~\cite{clip}, have demonstrated impressive generalization capabilities trained on internet-scale image-language pairs.
Initial works have shown that VLMs can enable 3D scene understanding towards \emph{open-vocabulary} recognition,
and offer capabilities to understand affordances, materials, activities, and properties of unseen environments without the need to train on costly 3D labeled data.
The development of such frameworks enables intelligent agents to perform arbitrary and complex tasks in novel environments.
\begin{figure}[t!]
  \centering
    \includegraphics[width=\linewidth]{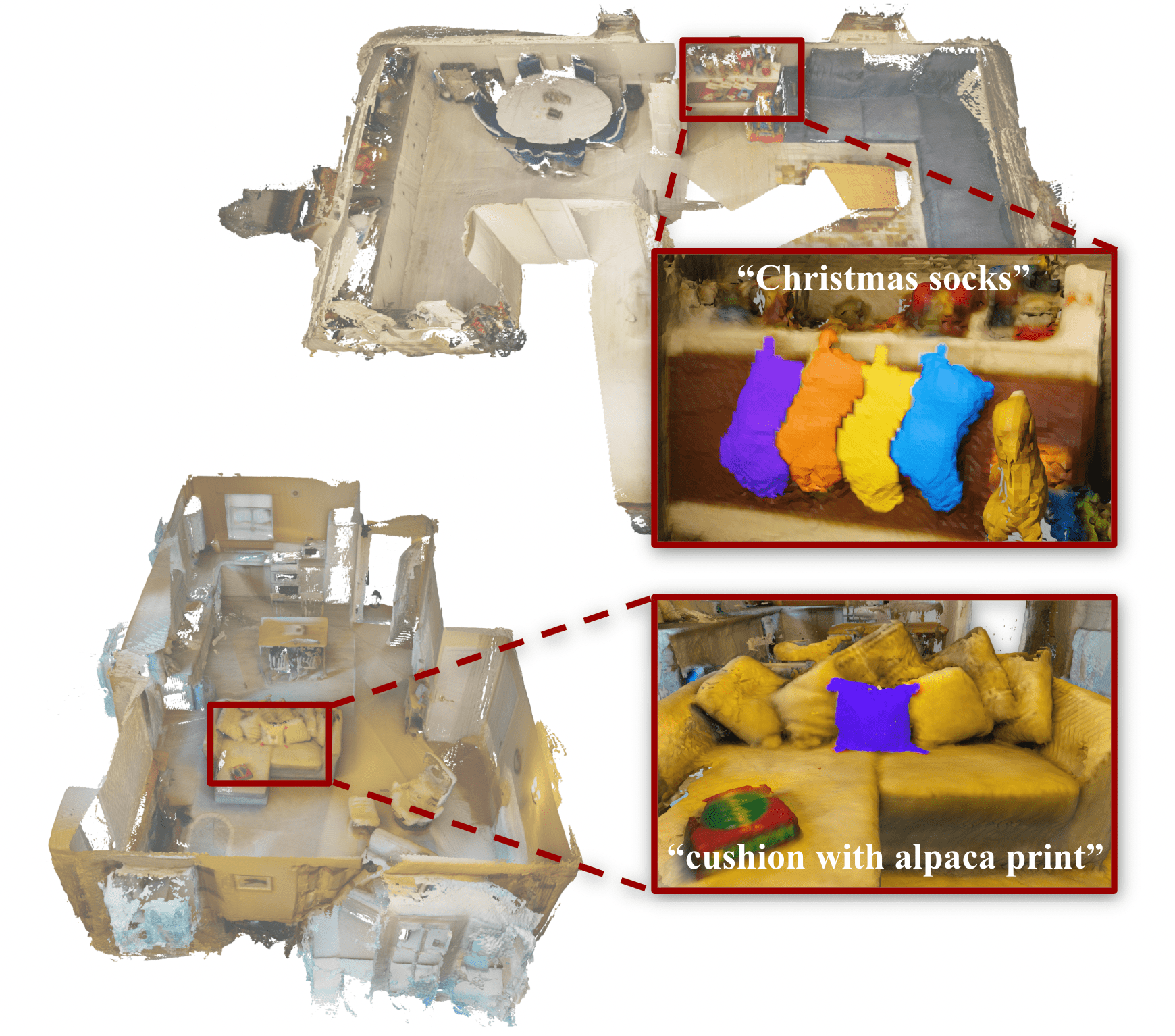}
    \vspace{-20pt}
    \caption{\textbf{Challenge overview.}
    Given a 3D scene in the form of a point cloud and associated posed images,
    the task is to segment all object instances described by a given free-form text query.}
  \label{fig:teaser}
  \vspace{-15pt}
\end{figure}
Driven by these motivations, the number of open-vocabulary 3D scene understanding approaches has been growing rapidly.
On one hand, there are explicit point-cloud based approaches such as OpenScene~\cite{Peng2022OpenScene}, PLA~\cite{Ding2022PLALO}, CLIP-FO3D~\cite{Zhang2023CLIPFO3DLF} and ConceptFusion~\cite{conceptfusion} which aim to create scene representations that allow querying the scene based on arbitrary open-vocabulary descriptions.
On the other hand, there are works such as LERF~\cite{Kerr2023LERFLE}, OpenNeRF~\cite{opennerf} and DFF~\cite{kobayashi2022distilledfeaturefields} that embed open-vocabulary features on an implicit scene representation.
Methods focusing on 3D instance segmentation include OpenMask3D~\cite{takmaz2023openmask3d},
OpenIns3D~\cite{Huang2023OpenIns3DSA} and Open3DIS~\cite{Nguyen2023Open3DISO3}.

There has also been a growing interest in creating a 3D scene graph representation \cite{sarkar2023sgaligner,Armeni_2019_ICCV} that allows for open-vocabulary querying such as ConceptGraphs~\cite{Gu2023ConceptGraphsO3}.
These open-vocabulary approaches go beyond the pre-defined capabilities of traditional 3D scene understanding methods operating in a closed-vocabulary setting. 

The goal of the workshop is to bundle these efforts towards 3D open-vocabulary scene understanding, and to discuss and establish clear task definitions, evaluation metrics, and benchmark datasets.

One of the important goals of this challenge is to help enable the quantitative comparison of open-vocabulary 3D scene segmentation methods.
In fact, due to the lack of benchmarks addressing various open-vocabulary concepts,
existing 3D open-vocabulary approaches mainly evaluate their capabilities qualitatively on individually selected queries and scenarios,
which differ from work to work.
The key goal of the workshop is to drive the research community inspire the research community to reason about the usability and generalization of their approaches in real-world scenarios.

In this iteration of the workshop, the challenge focuses on the open-vocabulary setting with arbitrary object descriptions,
which describe long-tail objects as well as object properties such as semantics, materials, affordances, and situational context.
In summary, the challenge is:

\begin{enumerate}
    \item \textbf{Task:} Given an open-vocabulary, text-based query, the aim is to localize and segment the object instances that fit best with the given prompt, which can describe object properties such as semantics, material type, affordances and situational context. 
    \item \textbf{Input:} An RGB-D image sequence and the corresponding 3D reconstructed geometry of a given scene, camera parameters (intrinsics, extrinsics), and an input query.
    \item \textbf{Output:} Instance segmentation masks in the point cloud that corresponds to the vertices of the provided 3D mesh reconstruction, segmenting the objects that fit best with the provided text prompt.
\end{enumerate}

In the following sections, we first introduce the challenge task, dataset and evaluation methodology (Sec.~\ref{sec:challenge}). Then, we share results from the top three winning teams, as well as the respective methods proposed by each team (Sec.~\ref{sec:methods}). 

\section{Challenge}
\label{sec:challenge}
In this section, we provide an overview of the challenge. More specifically, we describe the task, dataset, challenge phases and additional details including the evaluation.

\subsection{Task}
The task in this challenge is 3D open-vocabulary instance segmentation.
Given a 3D scene and an open-vocabulary, text-based query,
the goal is to segment all object instances that correspond to the specified query.
If there are multiple objects that fit the given prompt, each of these objects should be segmented, and labeled as separate instances.
The list of queries can refer to long-tail objects, or can include descriptions of object properties such as semantics, material type,
affordances and situational context. An example set of queries is provided in Tab.~\ref{tab:example_queries}.

\begin{table}[t!]
\small
    \centering
    \begin{tabular}{ll}
        \toprule
        No. & Example Queries \\
        \hline
        1 & cushion with an alpaca print \\
        2 & sofa right across the TV \\
        3 & Christmas stockings \\
        4 & present \\
        5 & wall art with a green background \\
        \bottomrule
    \end{tabular}
    \vspace{-5pt}
    \caption{Example queries from our challenge development set.}
    \label{tab:example_queries}
    \vspace{-15pt}
\end{table}

\subsection{Dataset}
Our challenge data is based on the existing ARKitScenes~\cite{baruch2021arkitscenes} dataset,
which is a large-scale indoor 3D dataset containing posed RGB-D image sequences,
3D mesh reconstructions from an iPad, as well as high-resolution Faro laser scans.
For the challenge, we use the RGB-D frames and the 3D reconstructions from the iPad.
While some existing datasets and benchmarks \cite{Chen2019ScanRefer3O, achlioptas2020referit_3d} are targeting language grounding in 3D scenes, these works primarily focus on identifying object relations or properties, whereas this challenge focuses specifically on long-tail object classes. 

\subsection{Challenge Phases}
To discourage overfitting to test data, the challenge consists of two phases: the \textit{development phase} and the \textit{test phase}.

\textbf{Development Phase:} In this first phase, the participants can download and use the whole \textit{training} split of the ARKitScenes dataset for their experiments. From these \textit{training} scenes, we annotate 5 example scenes for development purposes. More specifically, for each example scene, we first specify an open-vocabulary query, and then manually annotate objects corresponding to the given query, by segmenting them in the point cloud of the given scene. We refer to this subset as the \textit{challenge development set}.

\textbf{Test Phase:} In the second phase, the \textit{test phase}, we provide a subset of 25 scenes from the \textit{validation} split of the ARKitScenes dataset, we refer to this subset as the \textit{challenge test set}.
For each of these scenes, we provide an input text-query. The manually annotated ground truth segmentation masks corresponding to these queries remain hidden and the scores on this subset can only be computed by uploading the predictions to the online benchmark.

\subsection{Assets and Tools}
We rely on the \textit{raw} assets of the ARKitScenes dataset \cite{baruch2021arkitscenes}.
In particular, we rely on the depth images, RGB images, camera intrinsics, camera trajectories (poses), and the 3D reconstructed scene mesh -- these assets are ready available to to be downloaded from the ARKitScenes repository.
The specific assets we rely on for the challenge are listed below:

\begin{itemize}
    \item Reconstructed 3D mesh of the scene, generated by ARKit
    \item RGB images from the camera ($256 \times 192$) captured at $60$~Hz (\textit{low res. wide})
    \item RGB images from the wide camera ($1920 \times 1440$) captured at $10$~Hz (\textit{wide})
    \item Depth images acquired by Apple Depth LIDAR ($256 \times 192$) (\textit{low res. depth})
    \item Camera intrinsics for the low resolution camera
    \item Camera intrinsics for the high resolution camera
    \item Camera pose trajectory
\end{itemize}

We provide the challenge participants with instructions for downloading and pre-processing the data necessary for our challenge, as well as utility functions that are intended to guide the participants about how to load and use the provided data.
In particular, we provide the scene IDs for both the \textit{development} and \textit{test} sets,
as well as a list of text-queries associated with each scene.
For the development set, we also provide ground truth instance masks that correspond to the given text queries.
Finally, we provide an evaluation script and tools for the online benchmark submissions.

\subsection{Method Constraints}
As the primary goal of the workshop is to explore what open-vocabulary capabilities are available with the current methods and datasets, we in general do not enforce strict constraints regarding the participating methods, given the nature of open-vocabulary scene understanding task. The participants are allowed to benefit from pre-trained models such as CLIP~\cite{clip}. It is also allowed to pre-train on additional datasets other than the ARKitScenes \cite{baruch2021arkitscenes}.

\subsection{Submission Instructions}
Given an open-vocabulary text query, the participants are asked to segment all object instances that fit best with the given query. The expected result are object instance masks, and confidence scores for each mask.
We follow the same submission file structure and evaluation as ScanNet \cite{dai2017scannet}.

\subsection{Evaluation}
We follow the standard evaluation metrics for 3D  instance segmentation. For our challenge, in which the aim is to identify object instances that correspond to a given open-vocabulary query, we use the standard 3D instance segmentation metric Average Precision (AP). We compute AP$_{50}$ and AP$_{25}$ scores evaluated at IoU overlaps of $50\%$ and $25\%$ respectively. In addition, we report mAP scores obtained as the AP score averaged over the IoU overlap range $[0.5:0.95:0.05]$.

We provide evaluation scripts for all tasks and challenges which can be used by the participants to evaluate their algorithms on our challenge development set. For the challenge, we ask the participants to upload their predictions on the test scenes to the \href{https://eval.ai/web/challenges/challenge-page/2102/overview}{online challenge benchmark} page (hosted at eval.ai) for automatic evaluation. We have two separate benchmarks for the challenge development set and the challenge test set. The scores on the challenge test set can only be computed by uploading the predictions to our online benchmark. Final ranking of the teams is obtained by the results on the benchmark for the challenge test set. More specifically, we rank the teams based on the mAP score.

\subsection{Challenge Participation}
For our workshop challenge, 27 registered participants formed a total of 16 teams. Out of these 16 teams, 7 teams made submissions for the final test benchmark.

\subsection{Results}
Results of the top 3 winning teams of the workshop challenge are provided in Tab.~\ref{tab:challenge_results}.
While the performance of the first iteration of the workshop challenge are promising,
the absolute scores are still low for all submitted methods, which highlights the remaining challenges of open-vocabulary 3D scene understanding.
\begin{table}[h]
\small
    \centering
    \begin{tabular}{llcccc}
        \toprule
        Rank & Team & mAP (↑) & AP$_{50}$ (↑) & AP$_{25}$ (↑) \\
        \hline
        1 & PICO-MR & \textbf{6.08} & \textbf{14.08} & 17.67 \\
        2 & VinAI-3DIS & 4.13 & 12.14 & \textbf{39.41} \\
        3 & CRP & 2.67 & 5.06 & 13.98 \\
        \bottomrule
    \end{tabular}
        \vspace{-5pt}
    \caption{Top 3 winning teams from the workshop challenge.}
    \label{tab:challenge_results}
\end{table}

\begin{figure*}[!ht]
\centering
\vspace{25px}
\includegraphics[width=\linewidth]{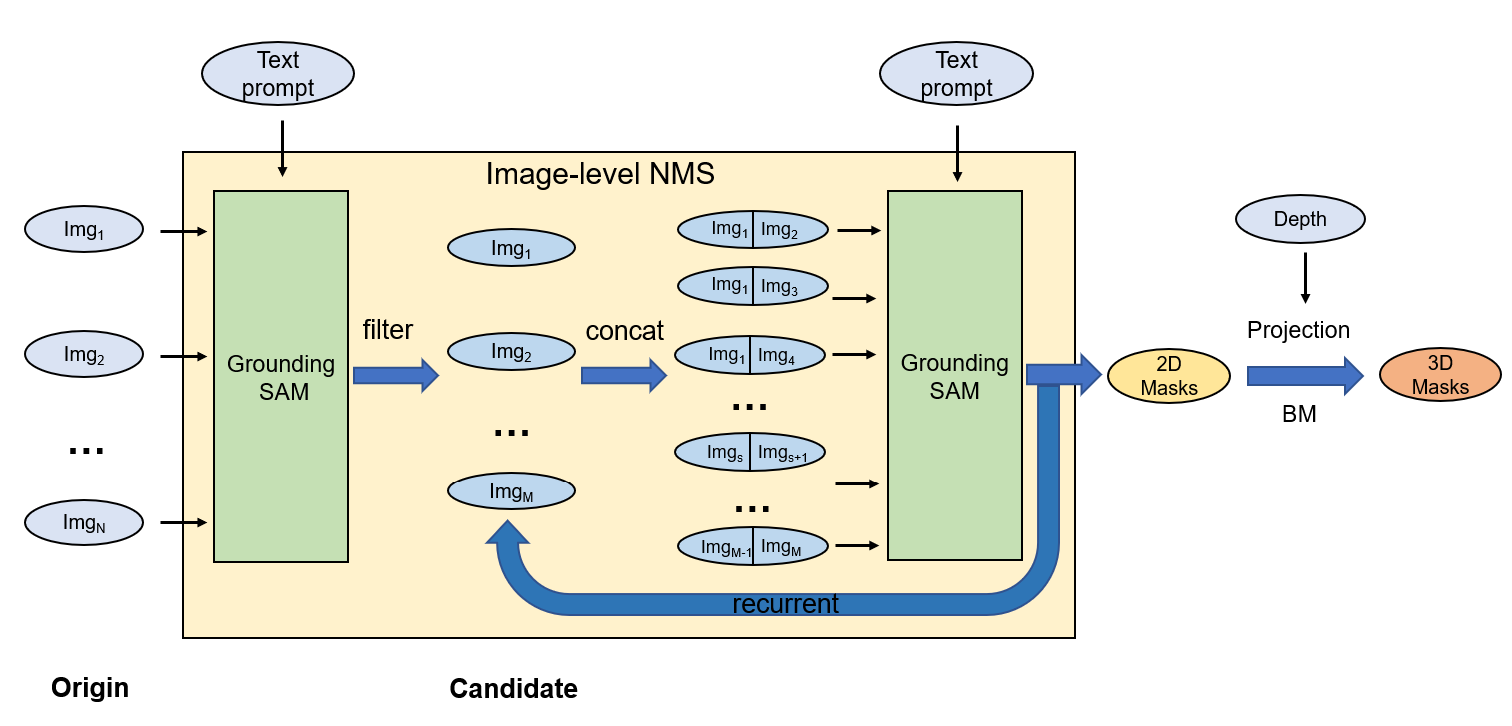}
\vspace{-1em}
\caption{
\textbf{Overview of the method proposed by the PICO-MR team.}  
An image-level NMS method based on Grounding SAM is designed to suppress the false positives generated by original Grounding SAM. 
Bidirectional Merging (BM) is the post process of SAM3D which iteratively merges adjacent point clouds to final 3D masks.
}
\vspace{-1em}

\label{fig:pico-mr-pipeline}
\end{figure*}

\section{Winning Methods}
\label{sec:methods}

In this section, the workshop challenge participants describe their winning methods.

\subsection{PICO-MR (1$^{st}$ Place)}
\emph{by Hongbo Tian\textsuperscript{1,2}, Chunjie Wang\textsuperscript{1}, Xiaosheng Yan\textsuperscript{1}, Bingwen Wang\textsuperscript{1}, Xuanyang Zhang\textsuperscript{1} and Xiao Liu\textsuperscript{1}.}
\emph{\footnotesize
$^1$PICO,~ByteDance \hspace{10px}
$^2$Beijing University of P\&T
}

\paragraph{Method.} We employ Grounding SAM \cite{liu2023grounding, kirillov2023segany} as our main component to find 2D masks in images. Due to the partial shoot of indoor scenes in the ARKitScenes \cite{baruch2021arkitscenes} dataset, many different objects look similar without context, which causes the open-set detection model to yield more false positives. 
To address this issue, we designed an image-level non-maximum suppression (NMS) method to suppress excessive false positives. Specifically, we first use Grounding SAM \cite{liu2023grounding} to generate candidate masks for target objects (\ie{}, any mask confidence $>$ 0.5), and then filter the remaining images without masks. Then we pair up all images with candidate masks two-by-two with horizontal concatenation, and feed these concatenated images to the Grounding SAM \cite{liu2023grounding} model again, which may result in three situations:
\begin{itemize}[leftmargin=5.5mm]
    \item [(1)] One part of the paired image contains valid masks, and the other does not.
    \item [(2)] Both have valid masks.
    \item [(3)] Neither has valid masks.
\end{itemize}

We keep the image parts with detected masks and filter the image parts without masks, which is essentially like an image-level NMS. We repeat the process above with the remaining images until all image parts contain valid masks. Finally, we use the Bidirectional Merging (BM) method \cite{yang2023sam3d} to merge all of the remaining 2D masks into 3D instance masks. In addition, to avoid the projection error caused by pose deviation of each frame, we first apply erosion on 2D masks before the BM stage. An overview of the whole pipeline is shown in Fig.~\ref{fig:pico-mr-pipeline}.

\subsection{VinAI-3DIS (2$^{nd}$ Place)}
\vspace{-2px}
\emph{by Phuc Nguyen\textsuperscript{1}, Khoi Nguyen\textsuperscript{1}, Anh Tran\textsuperscript{1} and Cuong Pham\textsuperscript{1,2}.}\\
\emph{\footnotesize
$^{1}$VinAI Research \hspace{10px}
$^{2}$Posts \& Telecommunications Institute of Technology
}

\begin{figure*}[!th]
\vspace{3px}
\includegraphics[width=\linewidth]{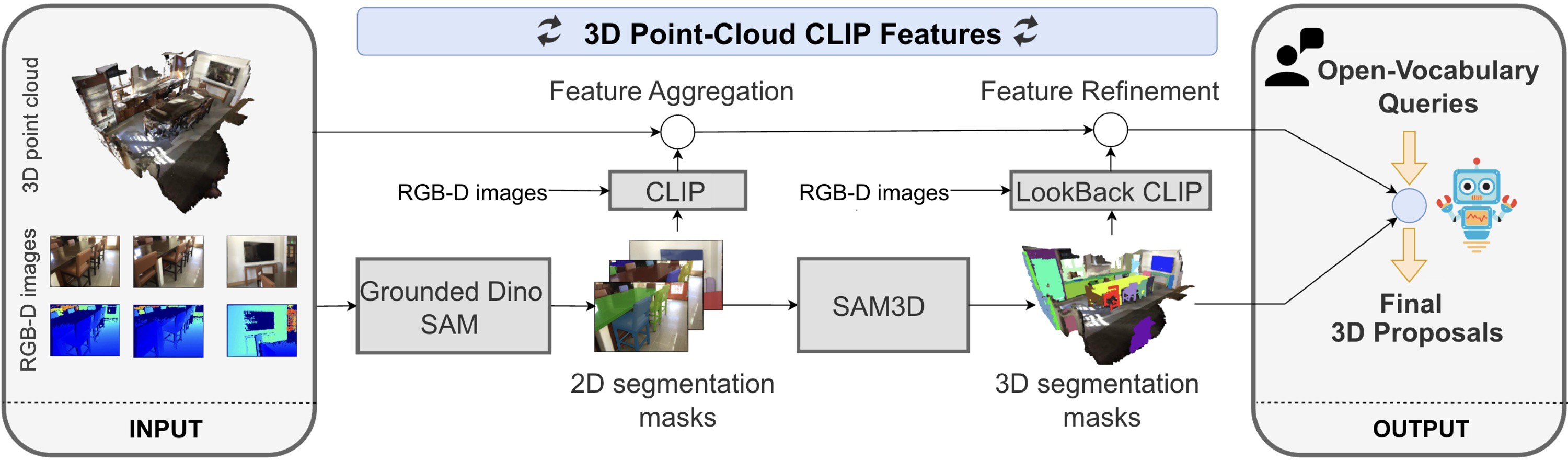}
\vspace{-5pt}
\captionof{figure}{\textbf{Overview of the method proposed by the VinAI-3DIS team.} }
\label{fig:vinai-3dis-pipeline}
\vspace{10pt}
\end{figure*}

\begin{figure*}[!th]
\vspace{5px}
\includegraphics[width=\linewidth]{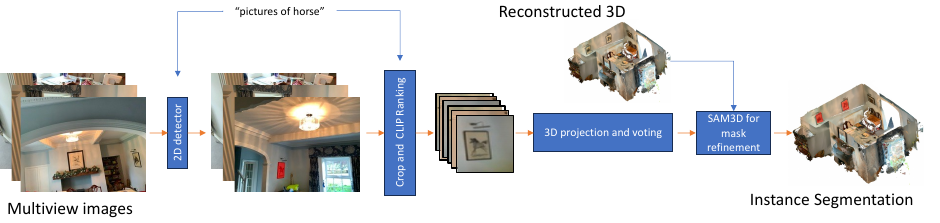}
\vspace{-20pt}
 \captionof{figure}{\textbf{Overview of the method proposed by the CLIP-ranked-projection (CRP) team.}}
\label{fig:crp-pipeline}
\vspace{-10px}
\end{figure*}

\paragraph{Method.} Our method processes both a 3D point cloud and an RGB-D sequence, resulting in the collection of 3D binary masks and refined 3D point cloud features. These components are subsequently used to identify target 3D proposals through an open-vocabulary query. We assume known camera parameters for each frame. The structure of our framework is depicted in Fig.~\ref{fig:vinai-3dis-pipeline}.

We leverage Grounding Dino \cite{liu2023grounding} and SAM \cite{kirillov2023segany} to generate class-agnostic 2D segmentation masks. These masks undergo an outer-cropping process before being passed through a CLIP \cite{clip} encoder to extract features. These features are then aggregated into 3D point-cloud CLIP features, using camera parameters and depth images for context. SAM3D \cite{yang2023sam3d} is subsequently applied to the generated 2D instance masks to produce elevated 3D segmentation masks.

To enhance the understanding of these masks, they are processed by the \textit{LookBack} CLIP module, where the masks are projected onto 2D images to generate multi-view outer-cropped CLIP features. These features play a crucial role in refining the previously obtained 3D point-cloud CLIP features.

During the inference stage, an open-vocabulary query is employed to identify the target 3D proposals. This is achieved by pooling the cosine similarities between the text embedding and the 3D point-cloud CLIP features for each 3D segmentation proposal.

\subsection{CRP: CLIP-ranked-projection (3$^{rd}$ Place)}

\emph{by Zhening Huang\textsuperscript{1}, Xiaoyang Wu\textsuperscript{2}, Xi Chen\textsuperscript{2}, Hengshuang Zhao\textsuperscript{2}, Lei Zhu\textsuperscript{3} and Joan Lasenby\textsuperscript{1}.} \\
\emph{{\footnotesize
$^{1}$University of Cambridge \hspace{10px}
$^{2}$HKU \hspace{10px}
$^{3}$HKUST (Guangzhou)
}}

\paragraph{Method.}
The inputs of our method include 2D images, depth maps, corresponding camera parameters, and 3D point cloud data. First, the open vocabulary text queries and 2D images are fed into a 2D detection model to extract instance masks in 2D images. Then, the detected masks are cropped out as small images and ranked with a CLIP \cite{clip} model by its similarity score with the input text queries. This design is used to filter out false-positive detections. The top N cropped images are selected as the potential target objects.
Once the masks in 2D are detected, we project them back to 3D space with the corresponding depth and camera parameters. We then conduct another layer of filtering by only keeping masks projected from more than two cropped images.
An overview of the proposed method is provided in Fig.~\ref{fig:crp-pipeline}.

\paragraph{Limitations.} While the 2D detector and CLIP-based ranking can accurately select the target region and mask location, the challenge remains in proposing good quality masks. SAM3D \cite{yang2023sam3d} did not perform well on the ARKitScenes \cite{baruch2021arkitscenes} dataset, leading to low-quality masks in the final output.

\section{Conclusion and Outlook}
This report summarizes the challenge of the first edition of
the OpenSUN3D 2023 Workshop on Open-Vocabulary 3D Scene Understanding held in conjunction with ICCV 2023 in Paris.
We have provided an overview of the challenge and the methods proposed by the winning teams of the competition.
We believe that the community will benefit from the proposed task and benchmark for the 3D open-vocabulary instance segmentation task.
The challenge data is already adapted by the research community as an independent datasets for evaluating open-vocabulary 3D instance segmentation \cite{Nguyen2023Open3DISO3}. In future iterations of the workshop, we would like to build on the results of this first iteration and offer more extensive evaluation datasets, including additional tasks that go beyond traditional object detection and instance segmentation tasks.

\paragraph{Acknowledgements.}
This work was partially supported by an ETH Career Seed Award funded through the ETH Zurich Foundation.
\newpage
{
    \small
    \bibliographystyle{ieeenat_fullname}
    \bibliography{main}
}

\end{document}